\pdfoutput=1

\def\paperformat{corl}

\def\formatieee{ieee}

\PassOptionsToPackage{table}{xcolor}

\ifx\paperformat\formatieee
    \documentclass[conference,letterpaper,10pt]{IEEEtran}
\else
    \documentclass{article}
    \PassOptionsToPackage{dvipsnames,svgnames,x11names,table}{xcolor}
    \usepackage{xcolor}
    \definecolor{mydarkblue}{rgb}{0,0.08,0.45}
    \usepackage{needspace}
    \PassOptionsToPackage{sort&compress}{natbib}
    \usepackage[final]{corl_2026}    %
    \hypersetup{
      pdftitle={HLO-AMP: Hierarchical Loco-Manipulation from Object-Aware Motion Priors},
      pdfauthor={Jean-Pierre Sleiman, Bilal Hammoud, He Li, Jonathan Foster, Sylvain Bertrand, Jessica K. Hodgins, Farbod Farshidian}
    }
    \makeatletter
    \renewcommand{\@notice}{%
        \enlargethispage{2\baselineskip}%
        \@float{noticebox}[b]%
            \vspace*{\baselineskip}%
            \footnotesize\@noticestring%
        \end@float%
    }
    \makeatother
\fi

\usepackage{amsmath}
\usepackage{amssymb}
\usepackage{bm}
\usepackage{booktabs}
\usepackage{float}
\usepackage{xcolor}
\usepackage{graphicx}
\graphicspath{{figures/}}
\usepackage{tikz}
\usetikzlibrary{arrows.meta, positioning, shapes.geometric, fit, backgrounds, calc, decorations.pathreplacing}
\usepackage{microtype}
\microtypesetup{factor=1100,stretch=20,shrink=20}
\usepackage[font=small,labelfont=bf,skip=7pt]{caption}
\renewcommand{\arraystretch}{1.0}
\newcommand{\tablefont}{\fontsize{8}{9.6}\selectfont}

\ifx\paperformat\formatieee\else
    \usepackage{enumitem}
    \setlist{nosep,topsep=2pt,partopsep=0pt,leftmargin=*}
    \renewcommand{\paragraph}[1]{%
        \par\smallskip\noindent\textbf{#1.}\hspace{0.5em}\ignorespaces}
\fi

\newcommand{\makecell}[1]{\begin{tabular}[c]{@{}c@{}}#1\end{tabular}}

\makeatletter
\newif\if@safefigfound
\newcommand{\safeincludegraphics}[2][]{%
    \begingroup
    \@safefigfoundfalse
    \IfFileExists{#2}{\@safefigfoundtrue}{%
        \IfFileExists{figures/#2}{\@safefigfoundtrue}{}%
    }%
    \if@safefigfound
        \includegraphics[#1]{#2}%
    \else
        \fbox{\parbox[c][0.18\textheight][c]{0.95\linewidth}{\centering Missing figure: \texttt{\detokenize{#2}}}}%
    \fi
    \endgroup
}
\makeatother

\ifx\paperformat\formatieee
    \usepackage[numbers,sort&compress]{natbib}
\fi

\ifx\paperformat\formatieee
    \providecommand{\keywords}[1]{%
        \begin{IEEEkeywords}#1\end{IEEEkeywords}}
    
    \usepackage{etoolbox}
    \usepackage{adjustbox}
    \AtBeginEnvironment{tabular}{\scriptsize\setlength{\tabcolsep}{3pt}}
    \BeforeBeginEnvironment{tabular}{\begin{adjustbox}{max width=\linewidth}}
    \AfterEndEnvironment{tabular}{\end{adjustbox}}
    \usepackage[colorlinks=true,linkcolor=blue,citecolor=blue,urlcolor=blue]{hyperref}
\fi

\title{HLO-AMP: \textbf{H}ierarchical \textbf{L}oco-Manipulation from \textbf{O}bject-\textbf{A}ware \textbf{M}otion \textbf{P}riors}

\ifx\paperformat\formatieee
    \author{%
        \IEEEauthorblockN{Anonymous Author(s)}\\
        \IEEEauthorblockA{%
            }%
    }
\else
    \author{
      Jean-Pierre Sleiman\quad
      Bilal Hammoud\quad
      He Li\quad
      Jonathan Foster\\
      \textbf{Sylvain Bertrand\quad
      Jessica K.\ Hodgins\quad
      Farbod Farshidian}\\
      Robotics and AI Institute, Cambridge, MA, USA
    }
\fi

\begin{document}
\raggedbottom
\maketitle
\pagestyle{plain}

\begin{abstract}
Hierarchical pipelines for humanoid loco-manipulation typically pair a
task-agnostic whole-body controller (WBC) with a downstream task policy.
Such WBCs are often trained through manual reward shaping, which
yields visibly machine-like motions; more consequentially, they are
trained without the manipulated object in the scene. Object dynamics are instead approximated through randomized end-effector force disturbances, a proxy that fails to reproduce the distributed multi-contact wrench a real payload applies across the arms and torso, with the gap becoming critical near the platform's payload limit. In this work, we train the low-level (LL) controller itself as an object-aware motion prior: a single policy trained via motion imitation on retargeted motion-capture (MoCap) references that span the core motion primitives of the downstream task, with the manipulated object physically present in the scene. This single stage inherits natural human-like coordination from the references and absorbs the realistic contact wrench at its source, while exposing the same compact task-space command interface used by command-tracking WBCs. A
reference-free high-level (HL) policy is then trained on top, with its
action channel anchored to the empirical envelope of the low-level
training data so that it generalizes the priors to a long-horizon
multi-stage task while structurally preventing queries that would push
the low level off-distribution. We instantiate the framework on a
Unitree~G1 for bimanual box pick-and-place: the same deployed policy 
generalizes across object masses and dimensions, initial and goal poses, 
and reactively recovers from intentional drops mid-carry.
Video: \url{https://youtu.be/viR_ULnRDQQ}.
\end{abstract}


\section{Introduction}
\label{sec:intro}

General-purpose humanoid platforms are designed to operate in
human-centric environments and physically interact with the objects
in them. Realizing
this promise requires whole-body loco-manipulation, in which a
control policy must coordinate locomotion, balance, posture, and
contact-rich object handling. When the robot
manipulates a payload, especially in dynamic rather than quasi-static
tasks where the object's inertial effects matter, the payload alters
the dynamics that govern balance.  Locomotion and manipulation are
therefore not separable subproblems that can be solved independently
and combined afterward.  These couplings make whole-body
loco-manipulation a central testbed for embodied intelligence on
humanoid platforms~\citep{HumanoidBoxLoco2023, HierBoxLoco2023,
ProHOI2026, HDMI2025, PhysHSI2025, VisualMimic2025, ResMimic2025,
ULC2025, WoCoCo2024, VIRAL2025, FALCON2025, TWIST2025, AMO2025}. In this paper, we study one representative and challenging
instance: bimanual box transport, in which the robot must approach a box, lift it with two hands, carry it, and place it at a desired location. This task requires the controller to regulate balance under varying inertial loads, sustain bilateral contact over long horizons, and transition smoothly between free locomotion and forceful interaction. Recent learning-based work approaches this problem along two complementary lines: One trains end-to-end whole-body policies to imitate time-indexed retargeted references~\citep{HDMI2025, OmniRetarget2025, ResMimic2025, Opt2Skill2024}: the imitation objective implicitly regularizes the policy toward natural-looking motion and removes the need for per-task reward shaping, but behaviors are limited to the reference distribution. The other factorizes the system into a task-agnostic command-tracking WBC~\citep{HOMIE2025, FALCON2025, ULC2025} with a higher-level task policy stacked on top~\citep{VIRAL2025, DoorMan2025, DemoHLM2025}. Such WBCs generalize across a wide command space and admit reuse across tasks, but they are produced through extensive manual reward shaping that yields visibly machine-like motion, and are trained without the manipulated object in the scene. Object dynamics are instead approximated through randomized end-effector force surrogates~\citep{FALCON2025}, which cannot accurately capture the distributed multi-contact wrench a real payload applies across the arms and torso.

In this work, we present a hierarchical reinforcement learning (RL) framework for humanoid loco-manipulation that addresses contact-rich complexity in a bottom-up manner, beginning at the low-level training stage. The LL policy is trained via motion imitation on retargeted MoCap with the object physically in the scene, absorbing the realistic multi-contact wrench at its source and inheriting natural human-like coordination without manual reward shaping. A reference-free HL policy then steers this frozen prior from only a sparse user-supplied goal. We formulate the HL policy's Markov decision process (MDP) so that its action channel structurally constrains commands to those realizable by the LL policy, avoiding off-distribution inputs that could degrade the frozen prior.  We instantiate the framework on a Unitree~G1 for bimanual box pick-and-place. A single deployed policy executes the full task with natural-looking whole-body motion, generalizes across box geometries and payload masses up to 7\,kg (approaching the platform's actuation envelope), and reactively recovers from external disturbances such as intentional mid-carry drops. Complementary simulation studies characterize the performance and robustness of our method, compare against end-to-end and hierarchical baselines under similar training conditions, and ablate each core component. We rely on MoCap for object-state feedback and leave on-board perception to future work. The priority here is \emph{generality, naturalness, and reliability} of the whole-body skill before integrating a perception stack.

\section{Related Work}
\label{sec:related}

\paragraph{Learning-based whole-body control for humanoids}
\label{sec:related:wbc}
WBCs for humanoids admit several concrete realizations, differing along two axes: the supervision signal, and whether the controller is holistic or decoupled into upper- and lower-body pairs. A first group is built around imitation of large-scale offline motion datasets. For instance, ExBody~\citep{Cheng2024ExBody} adopts a decoupled architecture (upper body tracks reference motion, lower body follows velocity commands), while OmniH2O~\citep{He2024OmniH2O}, TWIST2~\citep{TWIST22025}, and CLONE~\citep{CLONE2025} train holistic whole-body trackers over full-body kinematic references through teacher--student distillation. A second group avoids motion data entirely and is trained through manual reward shaping: HOMIE~\citep{HOMIE2025} couples height tracking with an upper-body pose curriculum in a decoupled architecture, FALCON~\citep{FALCON2025} factorizes control into a lower-body locomotion agent and an upper-body manipulation agent jointly trained with a force curriculum on randomized end-effector disturbances, and ULC~\citep{ULC2025} argues against hierarchical decompositions, training a single fine-grained policy that tracks root-level commands, and arm joint positions. Common to these systems, training either ignores the manipulated-object dynamics or uses randomized force disturbances as a surrogate~\cite{FALCON2025}.

\begin{figure*}[t]
    \centering
    \includegraphics[width=\linewidth]{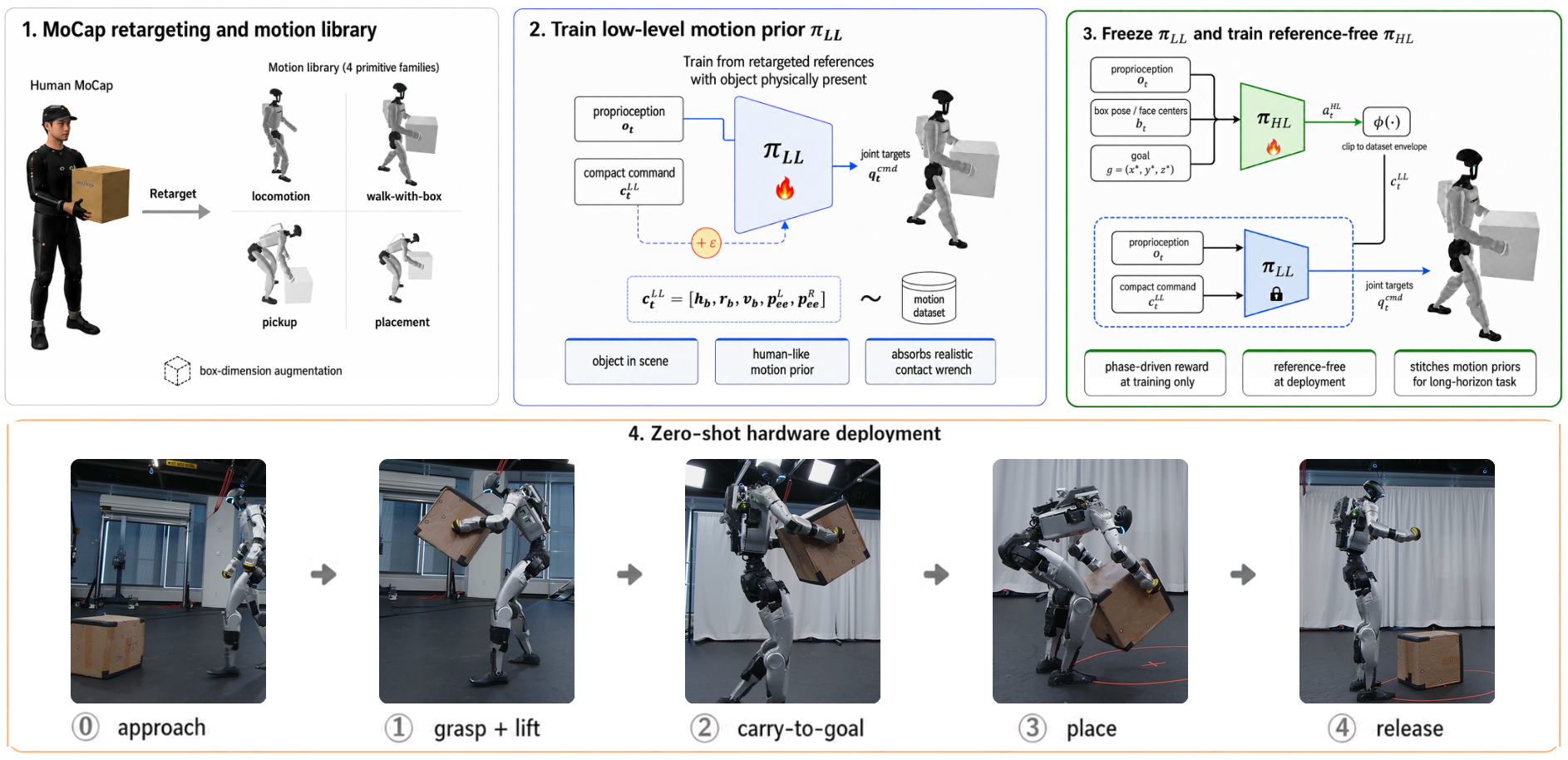}
\caption{\textbf{System Overview.} Retargeted MoCap data is used to build a compact motion library (locomotion, walking-with-box, pickup, placement). A low-level motion-imitation policy $\boldsymbol{\pi}_{\mathrm{LL}}$ is then trained as an object-in-scene prior, conditioned on a noisy task-space command. The frozen $\boldsymbol{\pi}_{\mathrm{LL}}$ is reused by a reference-free high-level policy $\boldsymbol{\pi}_{\mathrm{HL}}$ observing proprioception, box face centers, and a sparse goal, producing commands within the LL dataset envelope and stitching the learned priors into the full behavior for zero-shot deployment.}
\label{fig:overview}
\end{figure*}
\paragraph{Reusable skill spaces and motion priors} A separate line, originating in physics-based character animation, learns compact latent representations of diverse whole-body behaviors from which downstream policies sample to accelerate task learning, ranging from distilled motor-primitive embeddings~\citep{NPMP2018, CatchCarry2019, UHMR2023} to adversarially learned skill spaces~\citep{Peng2022ASE}; the recipe has been carried to real legged platforms as well~\citep{ImitateRepurpose2022, R2S2_2025}. As with the WBCs above, the resulting priors do not encode manipulation-induced dynamics, and they additionally introduce a non-trivial and hard-to-interpret latent-to-task conditioning interface that object-aware high-level modules must operate through.

\paragraph{Learning-based legged loco-manipulation} Legged loco-manipulation has recently been dominated by learning-based approaches, varying along two axes: how the long-horizon objective is shaped, and whether a reusable LL controller is exploited. A first cluster relies on reward shaping: Dao et al.~\cite{HumanoidBoxLoco2023} train one policy per stage and stitch them through a behavior graph, while WoCoCo~\citep{WoCoCo2024} avoids per-stage switching by decomposing long-horizon tasks into sequential contact stages within a single policy. These methods avoid reliance on motion data, but at the cost of meticulous reward engineering, and rarely constrain stylistic quality. A second cluster eliminates much of the reward-engineering burden by tracking pre-computed references from retargeted human MoCap~\citep{OmniRetarget2025,ResMimic2025}, video-captured human motion~\citep{HDMI2025}, or trajectory optimization~\citep{Opt2Skill2024,GuidedRLMultiContact2024}, at the cost of being limited to the reference distribution. A third cluster generalizes beyond reference distributions by stacking a higher-level module on a frozen RL whole-body controller. On humanoids, VIRAL~\citep{VIRAL2025} and DoorMan~\citep{DoorMan2025} train high-level policies on pretrained WBCs through teacher--student distillation, while DemoHLM~\citep{DemoHLM2025} relies on demonstration-conditioned behavior cloning; on quadrupedal mobile manipulators, a frozen WBC is reused for large-object non-prehensile manipulation through a sampling-based model-predictive-control (MPC) planner~\cite{Sumo2025} or a constrained-RL high-level policy on top~\cite{Dadiotis2025DynamicPushing}. These hierarchical pipelines can be used broadly but inherit the LL limitations identified above. Bridging work~\citep{ProHOI2026, PhysHSI2025, VisualMimic2025} retains motion-prior data while aiming to generalize beyond the reference distribution. PhysHSI~\citep{PhysHSI2025} relies on adversarial motion priors~\citep{AMP} to encourage stylized motion without strict tracking; VisualMimic~\citep{VisualMimic2025} employs teacher--student distillation to train a data-driven RL motion tracker with a task-specific downstream motion generator stacked on top; and Pro-HOI~\citep{ProHOI2026} adopts an imitation-based low-level policy that uses only the root pose and contact states as reference and relies on a kinematic motion planner to supply the root trajectory required by the task.

\paragraph{Positioning of our work} Our LL shares the object-tracking objectives of the reference trackers above~\citep{HDMI2025, ResMimic2025, OmniRetarget2025}, but is trained as a reusable prior for a reference-free HL policy rather than as a reference tracker, which requires the LL training adaptations of Sec.~\ref{sec:method:ll}. Among the bridging approaches, VisualMimic~\citep{VisualMimic2025} is the closest, as it also bounds the commands that its downstream motion generator sends to the tracker; however, it relies on a two-stage distilled keypoint tracker, whereas our prior is a single-stage RL policy with a compact WBC-style command. Moreover, beyond bounding the HL commands, we shape the HL objective to keep its behavior within the demonstrated motion manifold (Sec.~\ref{sec:method:hl}), and we target a multi-stage transport task involving large sustained interaction forces rather than individual skills. PhysHSI~\citep{PhysHSI2025} likewise requires no references at deployment, but its reported results are limited to light boxes and do not include recovery from large disturbances. Our aim is a single deployed policy that combines natural motion, generalization across object masses and dimensions, operation near the platform's torque envelope, and reactive recovery from large disturbances.

\section{Methodology}
\label{sec:method}

\subsection{Problem Formulation}
\label{sec:method:formulation}
\label{sec:method:problem}

As depicted in Figure~\ref{fig:overview}, we cast humanoid loco-manipulation as a hierarchical RL problem decomposed into a low-level whole-body tracking policy $\pi^{\mathrm{LL}}$ and a high-level loco-manipulation policy $\pi^{\mathrm{HL}}$, both trained in a simulation environment containing the manipulated object in the scene.
The LL policy $\pi^{\mathrm{LL}}(\mathbf{a}^{\mathrm{LL}}_t \mid \mathbf{o}^{\mathrm{LL}}_t, \mathbf{c}^{\mathrm{LL}}_t)$ maps a proprioceptive observation $\mathbf{o}^{\mathrm{LL}}_t$ and a compact task-space command $\mathbf{c}^{\mathrm{LL}}_t$ to a joint-level target $\mathbf{a}^{\mathrm{LL}}_t$. The HL policy $\pi^{\mathrm{HL}}(\mathbf{a}^{\mathrm{HL}}_t \mid \mathbf{o}^{\mathrm{HL}}_t, \mathbf{g})$ extends proprioception with object-localizing measurements and a sparse user-supplied goal $\mathbf{g}$ specifying the target object configuration in the robot frame. Its action $\mathbf{a}^{\mathrm{HL}}_t$ is mapped to the LL command through a transformation $\mathbf{c}^{\mathrm{LL}}_t = \phi(\mathbf{a}^{\mathrm{HL}}_t)$ designed to keep inputs to $\pi^{\mathrm{LL}}$ within its training distribution.
The two policies are trained sequentially: $\pi^{\mathrm{LL}}$ is first trained as a motion-imitation controller on retargeted human-motion references, then frozen and reused as the motion prior over which $\pi^{\mathrm{HL}}$ is trained. 

\subsection{Data Collection and Retargeting}
\label{sec:method:retargeting}

Reference clips are recorded via MoCap of a single human subject manipulating a box, with the box trajectory captured alongside the human motion. The seed library is organized into four primitive families (free locomotion, walking-with-box, box pickup, and box placement). We retarget the clips onto the Unitree~G1 via the interaction-mesh formulation~\citep{ho2010interaction, OmniRetarget2025}, which preserves the relative geometry of the agent--object--terrain system rather than matching joints one-to-one. To bridge the human/G1 anthropometry gap, we apply per-segment scaling, chain-rebuild, regrounding, and grasp-contact-preservation corrections. For each manipulation primitive, we augment the dataset to 10 box variants, comprising three base sizes (small, medium, large) and seven rotational variations, each re-computed via the same interaction-preserving optimization to obtain contact-consistent references. Further details can be found in Appendix~\ref{app:retargeting}.

\subsection{Low-Level Whole-Body Tracking MDP}
\label{sec:ll_policy}
\label{sec:method:ll}

Our LL training follows \emph{ZEST}~\cite{ZEST2025}, a proprioception-only motion-imitation recipe that transfers zero-shot to hardware across diverse motion corpora on multiple legged platforms. We adapt it to serve as a frozen LL motion prior for downstream loco-manipulation. Each training episode contains a single primitive from the motion library above, and stitching them into the long-horizon task is deferred entirely to $\pi^{\text{HL}}$. The adaptations specific to our setting are described below.

\paragraph{Object-in-scene training}
We train $\pi^{\text{LL}}$ to jointly track robot and object references under the resulting interaction forces. Despite the box being in the scene, it is not observed by $\pi^{\text{LL}}$: the box pose is randomized during training of $\pi^{\text{HL}}$ and at deployment, so feeding it to $\pi^{\text{LL}}$ would push its observation off-distribution. We retain \emph{ZEST}'s standard tracking and regularizer terms and add a tracking term and an early-termination predicate on the box state, tying the imitation objective to the manipulated object without requiring it to be observed by the actor. We also adopt the same adaptive-RSI scheme, which biases reset selection toward time bins with higher failure rates; however, the randomization offset around the sampled state is kept small here, both to avoid spawning the robot inside the box and because resets far from the reference would lead to failed episodes.  

\paragraph{Compact WBC-style command interface}
Instead of conditioning the actor on the full next-step state reference used by \emph{ZEST}, $\pi^{\text{LL}}$ is conditioned on a compact task-space command $\mathbf{c}_t^{\text{LL}} = [\,h_b,\,\mathbf{r}_b,\,\mathbf{v}_b,\,\mathbf{p}_{ee}^{L},\,\mathbf{p}_{ee}^{R}\,] \in \mathbb{R}^{12}$, where $h_b$ is the base height, $\mathbf{r}_b$ the base orientation (roll, pitch), $\mathbf{v}_b$ the planar base twist, and $\mathbf{p}_{ee}^{L,R}$ the end-effector positions in the torso frame. We expose only end-effector positions and find this sufficient for the four primitive families considered; an orientation channel can be appended without changing the rest of the pipeline. Because $\mathbf{c}_t^{\text{LL}}$ will ultimately be produced by the stochastic HL policy rather than by clean retargeted references, we inject bounded noise on $\mathbf{c}_t^{\text{LL}}$ during training (Appendix~\ref{app:obs}); without it, $\pi^{\text{HL}}$ fails to reliably steer $\pi^{\text{LL}}$ once the two are composed.

\paragraph{Action parameterization and residual-removal curriculum}
The LL output $\mathbf{a}_t^{\text{LL}}$ is a per-DoF scaled residual added to a feedforward joint baseline. The natural baseline would be the reference joint positions $\hat{\mathbf{q}}_t$, but these are unavailable at deployment since $\pi^{\text{HL}}$ produces $\mathbf{c}_t^{\text{LL}}$ rather than full joint references. We therefore apply a curriculum
\begin{equation}
\mathbf{q}_t^{\text{cmd}} \;=\; (1-\beta_t)\,\hat{\mathbf{q}}_t \;+\; \beta_t\,\mathbf{q}_t \;+\; \boldsymbol{\Sigma}\,\mathbf{a}_t^{\text{LL}},\qquad \beta_t:0\to 1,
\label{eq:residual_curriculum}
\end{equation}
that slides the baseline from $\hat{\mathbf{q}}_t$ to the robot's own measured configuration $\mathbf{q}_t$ by the end of training; $\beta_t$ is scheduled via \emph{ZEST}'s per-bin failure-driven mechanism, repurposed from its assistive-wrench curriculum. This yields higher tracking fidelity and fewer early terminations than fixing the feedforward term to either the default standing pose or the measured configuration from the start (Tab.~\ref{tab:residual_curriculum}).

\subsection{High-Level Loco-Manipulation MDP}
\label{sec:hl}
\label{sec:method:hl}

We freeze $\pi^{\text{LL}}$ and train $\pi^{\text{HL}}$ on top to solve our downstream loco-manipulation task in settings that go beyond the LL's motion dataset. The details of the MDP formulation are described below.

\paragraph{Action space}
The HL policy network output is mapped to the LL command through
\begin{equation}
\mathbf{c}_t^{\text{LL}} \;=\; \phi(\mathbf{a}_t^{\text{HL}}) \;=\; \mathrm{clip}\!\left(\mathbf{c}^{\text{nom}} + \sigma\cdot\mathbf{a}_t^{\text{HL}},\;\mathbf{c}^{\text{ds}}_{\min},\,\mathbf{c}^{\text{ds}}_{\max}\right),
\label{eq:phi_hl}
\end{equation}
where $\mathbf{c}^{\text{nom}}$ is a nominal command corresponding to the robot's stance configuration extracted from the LL motion dataset, $\sigma$ is a manually-tuned scalar parameter that controls the exploration range around it, and $[\mathbf{c}^{\text{ds}}_{\min}, \mathbf{c}^{\text{ds}}_{\max}]$ is the per-dimension empirical envelope of the dataset. The per-dimension clip ensures that the HL policy starts in-distribution for the LL policy and remains there throughout training and deployment. To avoid task-specific tuning of $\mathbf{c}^{\text{nom}}$ and $\sigma$, both can be extracted directly from the LL reference dataset as the empirical mean and standard deviation of the command distribution; in our case, both options resulted in similar performance.

\paragraph{Observations}
The HL policy is reference-free at deployment. Its observation contains the same proprioceptive signals as the LL policy augmented with the previous HL action $\mathbf{a}_{t-1}^{\text{HL}}$, the box pose, and the user-supplied goal. The box pose representation is given by the relative 3D positions of its six face centers in the torso frame: this avoids orientation singularities, gives the HL policy direct access to contact-relevant geometry, and lets it learn its own low-dimensional representation of object geometry, orientation, proximity, and grasp affordance. The goal $\mathbf{g}=(x^\star,y^\star,z^\star)$ is the target box position, expressed in the same torso frame. Both the box face-center positions and the goal vector are hard-clipped per axis at every observation step; distances beyond the clip saturate to the same value, forcing the HL policy to depend on direction and on local geometry once the target enters the reachable regime. As a consequence, the deployed system tolerates arbitrarily large initial robot--box and box--goal offsets despite training under a much narrower randomization range. Because the MoCap system assigns box axes arbitrarily, the six face-center entries are randomly permuted once per episode at reset to decouple input-slot labeling from the physical box pose; orientation randomization would achieve the same invariance but conflicts with the contact-rich RSI states introduced below.

\paragraph{Phase-conditioned reward}
\label{sec:hl_reward}
To shape exploration over our long-horizon task, we drive training with a deterministic scheduler over five phases: (0)~\textsc{walk-to-box}, (1)~\textsc{lift-box}, (2)~\textsc{carry-to-goal}, (3)~\textsc{place-box} at $\mathbf{g}$, and an absorbing (4)~\textsc{idle}. Forward transitions are gated by predicates over physically meaningful quantities (position and heading tolerances, bilateral arm--box contact, and box height). Phase~2 additionally admits a dropped-box fallback that returns the scheduler to phase~0, while the placement predicate sends it to \textsc{idle}. The full transition function and thresholds are given in Appendix~\ref{app:planner}; the thresholds are read off the demonstrations (e.g., the approach distance from the base to the nearest box face), so a single set serves all box geometries without per-geometry tuning, and training is not highly sensitive to them as long as each keeps the next phase reachable. The total per-step reward $r_t = r_t^{\text{task}} + r_t^{\text{reg}} + r_t^{\text{alive}}$ combines a task reward, a regularization-and-limit penalty, and a constant alive bonus. The task reward is phase-based: the scheduler state $s_t$ selects which per-phase reward $R_s$ is active at each step,
\begin{equation}
r_t^{\text{task}} \;=\; w^{task} \cdot \sum_{s \in \mathcal{S}} \mathbf{1}_{\{s_t = s\}}\,R_s(\mathbf{x}_t,\, \dot{\mathbf{x}}_t,\, \mathbf{f}_t).
\label{eq:task_reward}
\end{equation}
Each $R_s$ is normalized to a maximum of $1$, preventing any single phase from dominating, and scores progress through a velocity projection $\hat{r} = \tanh(\alpha\,\mathbf{v}\cdot\hat{\mathbf{e}})$, which avoids length-scale-sensitive position-error terms (explicit per-phase forms $R_0$--$R_4$ in Appendix~\ref{app:reward_terms}). The velocity-projected signal is available from the first step the robot moves toward the target and does not vanish near the goal. The $\tanh$ also saturates the per-step reward, discouraging rush-to-goal behavior and keeping velocities within the natural range that the LL prior can track. The regularization term $r_t^{\text{reg}}$ retains the LL penalties and adds three HL-specific terms: a pre-clipped dataset-envelope penalty, an action-rate penalty, and a command-tracking penalty (Appendix~\ref{app:reward}). Crucially, the actor does \emph{not} observe the scheduler state $s_t$. This lets it deviate from the scheduler's strict ordering when a transition threshold is imperfect. It also removes any deployment-time dependence on contact-gated transitions, which would otherwise require a privileged contact stream. The scheduler is thus a training-time reward-shaping device only, and is not part of the deployed system.

\paragraph{Initialization \& termination}
We apply RSI on the HL side restricted to frames from the box-pickup demonstrations: with some probability the robot and box are jointly reset to a mid-pickup configuration, otherwise the robot is initialized far from the box and must discover the contact configuration through phase~0. This targeted RSI directly exposes the HL to the contact-rich phase-1 states that the staged reward alone is slow to discover; once the value function is well-shaped around these states, the learning signal propagates backward to the approach phase. We retain the LL's reference-independent early-terminations~\cite{ZEST2025} and add a stage-specific time-out: if the pickup phase is not completed within a fixed time budget, the episode is terminated. This prevents the policy from spending excessive time in uninformative configurations.

\subsection{Training Setup}
\label{sec:training}

Both the LL and HL policies are trained in IsaacLab~\cite{Mittal2025IsaacLab} with 8192 parallel environments at a shared control rate of 50\,Hz, and optimized with Proximal Policy Optimization (PPO)~\cite{Schulman2017PPO}. Network architectures and PPO hyperparameters are detailed in Appendix~\ref{app:training_setup}. Training spans three representative box geometries (small, medium, and large) with masses between $0.5$--$8.0$~kg. Both stages randomize the same set of factors to support sim-to-real transfer (e.g., robot and object mass, friction, PD gains). The HL additionally varies the box's source and target locations, and the robot's initial pose relative to the box. Source- and target-side support tables are spawned with asymmetric height and probability ranges; the exact ranges and rationale, together with all domain-randomization details, are given in Appendix~\ref{app:dr}. We adopt an asymmetric actor--critic scheme~\cite{Pinto2017AsymmetricAC} in both stages: the actor receives only the deployment-time observations described in Sec.~\ref{sec:method:ll} and Sec.~\ref{sec:hl}, while the critic additionally consumes privileged simulator states. Detailed observation lists are provided in Appendix~\ref{app:obs}.

\section{Experimental Results}
\label{sec:experiments}

We evaluate the framework on hardware to demonstrate zero-shot sim-to-real transfer, generalization across box geometries and masses, recovery from external disturbances, generalization to out-of-distribution (OOD) conditions, and the natural-looking motion quality inherited from LL training. We then characterize each design choice through baseline comparisons and targeted ablations in simulation, with a full robustness sweep deferred to Appendix~\ref{app:robustness}.

\subsection{Hardware Results}
\label{sec:hw}

All hardware experiments run on a Unitree~G1 with a single policy and no per-trial tuning. The actor receives onboard proprioception and the 6-DoF box pose from an external motion-capture system. Each condition was attempted three times, and all attempts succeeded except at 7 and 8~kg (detailed below). The supplementary video shows one representative take per condition chosen for visual quality rather than as the first successful trial.

\begin{figure}[t]
\centering
\safeincludegraphics[width=\columnwidth]{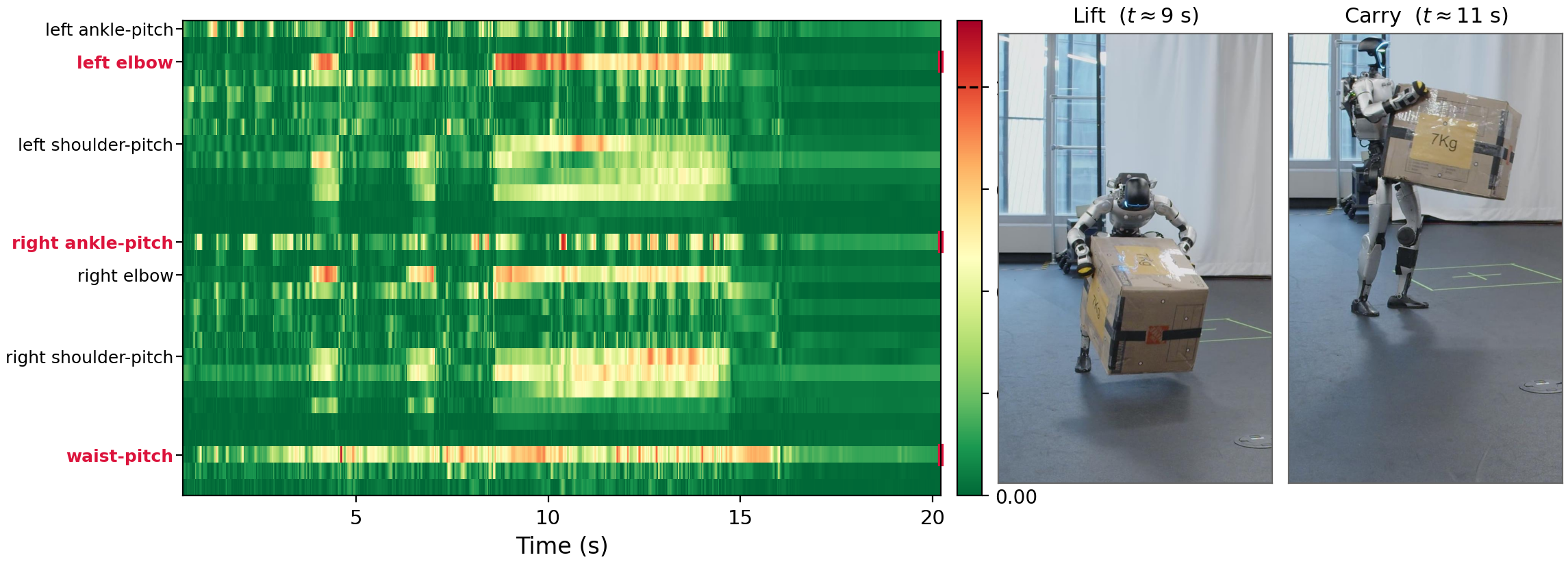}
\caption{Per-joint torque during the 7~kg experiment, normalized by each joint's effort limit (only discussed joints labeled; over-limit in bold red, dashed line marks 100\,\%). Snapshots: lift ($t\!\approx\!9$~s) and carry ($t\!\approx\!11$~s).}
\label{fig:torque_heatmap}
\end{figure}

\paragraph{Stress-test pick-and-place}
We probe the deployed policy along four dimensions of single-box pick-and-place: \emph{Baseline Operation.} With a $\approx 0.5$~kg box and a single ground goal, the policy completes the full approach--grasp--lift--carry--place sequence; the same transfers to a $\approx 2$~kg box. \emph{Disturbance recovery.} We then introduce two disturbances on each box: pushing it mid-carry until it drops from the robot's grasp, and displacing it from the goal after placement. In both cases, the controller successfully retries and recovers---re-grasping the dropped box in the first case and re-approaching and re-placing it in the second. \emph{Heavy payload.} Trials with $\approx 5$ and $\approx 7$~kg boxes succeed despite the policy having no access to box mass (domain randomization over object inertia covers this range), with the 7~kg trials (2/3 successful) requiring several lifting retries before the box is secured and inducing a few mid-sequence drops that the policy recovers from. The payload nonetheless drives the robot to the edge of its actuation envelope: the 7~kg torque heat-map (Fig.~\ref{fig:torque_heatmap}) shows three joints exceeding their effort limit during a sustained high-load epoch ($t\!\approx\!4$--$15$~s), with several more operating in the 80--99\,\% band (full breakdown in Appendix~\ref{app:torque_trace}). An $8$~kg box was also lifted but repeatedly dropped and never completed (0/3), with runs ending in thermal shutdown or a preemptive stop. The 7--8~kg trials sit at the edge of the training mass range and deliberately probe the actuation limit: brief effort-limit exceedances are feasible, but sustained operation there leads to thermal shutdown, and the dominant failure is grasp loss rather than carry instability. Under nominal payloads, the robot runs for long durations without such failures. \emph{OOD geometry.} A pick-and-place with a light $\approx 60$~cm cube---exceeding the largest training box ($54 \times 42 \times 39$~cm) along every axis---completes successfully. The face-center observation enables the HL to adapt its approach and grasp configurations to the larger geometry, indicating that the learned representation extrapolates reasonably beyond the training envelope.

\paragraph{Sequencing experiments}
In all subsequent experiments involving multiple boxes, we briefly switch to an idle policy before registering the next target box (the policy tracks one object at a time). \emph{Multiple goals.} With a medium $\approx 1$~kg box, four target locations are issued one at a time, with each new goal sent only after the box reaches the previous one. The policy completes all four segments, with every placement landing within $30$~cm of its target. \emph{Multiple boxes.} We then run the policy continuously across four boxes of different dimensions and masses, delivered back-to-back to the same goal. \emph{Mixed start and target heights.} Finally, we run two sequences, each starting from a small box stacked on top of a larger $\approx 2$~kg box (effective initial height $\approx 40$~cm); the target table stands at $75$~cm, the upper end of the training distribution. Since the policy observes only a single target box at a time, the supporting box underneath and the target table are both unobserved obstacles, and collisions during the lift or placement are handled as unstructured disturbances. In the first sequence, both box targets are on the table; while placing the larger box, a collision with the table edge occurs and the policy recovers, retrying at a higher carry height to complete the task. In the second sequence, the small box is placed on the table first and the larger box on the ground in front of it.

\paragraph{{Discussion}}
A single deployed policy handles every condition above. This is enabled by three components of our framework: object-in-scene LL training absorbs the contact-rich load coupling that a held box imposes on the arms and torso; the HL observation interface reduces unseen layouts to in-distribution inputs; and the phase-blind actor combined with the scheduler's drop fallback gives an endogenous recovery loop with no external trigger. The motion-imitation objective additionally endows the deployed behavior with natural-looking whole-body coordination, with smooth, quiet heel-toe foot transitions and a near-straight knee configuration. This contrasts with the flat-footed stomping and bent-knee crouch that reward-shaped humanoid controllers commonly converge to.

\subsection{Simulation-based Evaluation}
\label{sec:sim_perf}
\label{sec:sim_compare}
\label{sec:sim_ablation}
\label{sec:sim_robust}

We complement the hardware trials with large-scale evaluation in simulation across 10{,}000 parallel environments, under the same domain randomization applied during training; episodes are 20\,s long unless stated, and an episode succeeds if the box is placed within 0.50\,m of the goal (the training-time placement predicate, Appendix~\ref{app:planner}). We split success by payload into \emph{light} ($< 4$\,kg) and \emph{heavy} ($\geq 4$\,kg), and report the \emph{dropped-box rate} (fraction of episodes in which bilateral contact is lost at least once) and a \emph{naturalness-and-smoothness} (N\&S) \emph{index} averaging six lower-is-better motion-quality ratios relative to Ours (Appendix~\ref{app:ns_index}).

\paragraph{Performance, recovery, and OOD robustness}
Table~\ref{tab:sim_perf} reports task success as a function of episode budget. The policy completes 80.7\,\% of episodes at 20\,s, with success rising and plateauing past 30\,s and heavy boxes consistently trailing light. The dropped-box rate consistently exceeds the failure rate, so most 20\,s failures are episodes that would have completed given a longer recovery window; a per-phase breakdown (Appendix~\ref{app:phase}) shows that the carry phase accounts for the largest share of failures and that a 40\,s budget more than halves the carry-phase failures. Doubling the initial-pose and goal-placement ranges (last row) costs only completion time, not task success: with a 40\,s budget the policy still reaches 82.1\,\%, within 6 percentage points of the in-distribution result. This follows from the HL observation design, under which OOD box and goal positions saturate onto in-distribution inputs. A complementary three-axis stress test (Appendix~\ref{app:robustness}) further characterizes task-success degradation when box mass, external pushes, and box dimensions are each scaled well beyond the training distribution: the policy is \emph{least} sensitive to external pushes, where stronger pushes mainly increase the dropped-box rate rather than the failure rate; \emph{moderately} sensitive to payload mass, with the bottleneck being raw payload against the platform's actuation limits; and \emph{most} sensitive to box dimensions, since geometry changes the grasp and contact configuration that the face-center observation can only partially absorb.

\begin{table}[t]
\centering
\begin{minipage}[b]{0.425\columnwidth}
\caption{Task success (\%) of Ours vs.\ episode length; last row doubles init/goal ranges relative to training.}
\label{tab:sim_perf}
\end{minipage}\hfill
\begin{minipage}[b]{0.55\columnwidth}
\caption{Comparison to different baselines (20\,s). N\&S idx defined in the text: $1.00=$ Ours, higher is worse.}
\label{tab:sim_compare}
\end{minipage}\\
\tablefont
\setlength{\tabcolsep}{2.8pt}
\begin{minipage}[t]{0.425\columnwidth}
\centering
\rowcolors{2}{white}{gray!15}
\begin{tabular}{lcccc}
\toprule
Setting & \makecell{Succ.\\(all)} & \makecell{Succ.\\(light)} & \makecell{Succ.\\(heavy)} & \makecell{Dropped\\rate (\%)} \\
\midrule
20\,s & 80.7 & 85.8 & 70.5 & 26.8 \\
30\,s & 87.5 & 91.5 & 79.4 & 28.0 \\
40\,s & \textbf{87.9} & \textbf{91.9} & \textbf{80.2} & 28.4 \\
\midrule
40\,s, $2\times$ range & 82.1 & 87.5 & 71.5 & 33.2 \\
\bottomrule
\end{tabular}
\end{minipage}\hfill
\begin{minipage}[t]{0.55\columnwidth}
\centering
\rowcolors{2}{white}{gray!15}
\begin{tabular}{lccccc}
\toprule
Method & \makecell{Succ.\\(all)} & \makecell{Succ.\\(light)} & \makecell{Succ.\\(heavy)} & \makecell{Dropped\\rate (\%)} & \makecell{N\&S\\idx} \\
\midrule
Monolithic E2E        & 3.6  & 3.6  & 3.5  & 0.0  & $-$ \\
WBC (no object)       & 50.3 & 65.7 & 19.7 & 42.9 & 1.29 \\
WBC (force surrogate) & 77.1 & \textbf{88.4} & 55.1 & 78.0 & 1.16 \\
\textbf{Ours}         & \textbf{80.7} & 85.8 & \textbf{70.5} & \textbf{26.8} & \textbf{1.00} \\
\bottomrule
\end{tabular}
\end{minipage}
\end{table}

\paragraph{Comparison to baselines}
We compare against three alternatives (Table~\ref{tab:sim_compare}), all sharing our domain randomization, PPO configuration, compute budget, and task; the WBC baselines also use our command interface and HL training setup, so only the LL differs. \emph{Monolithic E2E} is a single end-to-end policy without the HL/LL decomposition or motion prior, sharing our HL's observations and task reward plus upright-torso terms during walking and carrying. The two \emph{WBC} baselines replace our LL with a command-tracking controller trained on a large human-mocap dataset of walking and reaching with both end-effectors (including the box loco-manipulation trajectories); neither sees the box during training, and the second adds an end-effector force surrogate that randomizes end-effector inertia and applies bounded force/torque disturbances~\cite{FALCON2025}. The monolithic policy effectively fails (3.6\,\%): it walks to the box (91.6\,\% reach the lift phase) but never discovers bimanual grasping, and its few ``successful'' trajectories kick the box along the ground until it reaches the goal. The object-free WBC reaches a moderate overall success, with a striking light/heavy split exposing the cause: a controller that never feels the payload wrench during training destabilizes under load. The force-surrogate variant recovers most of the overall success (77.1\,\%) but at a 78.0\,\% dropped-box rate. Ours achieves the highest overall success, the lowest dropped-box rate among the methods that actually transport the box, and the lowest N\&S index.

\needspace{5\baselineskip}
\paragraph{Component ablations}
Table~\ref{tab:sim_ablation} removes one component at a time. Every ablation collapses success from 80.7\,\% to 3--16\,\%, confirming no component is redundant. Removing observation noise or the dataset-envelope action clipping breaks the HL$\rightarrow$LL interface: the HL emits commands the LL was never trained to track, resulting in the highest OOD-command frequency and N\&S index. Removing the object from LL training (even with an EE-force surrogate) or removing the targeted HL RSI leaves locomotion intact but the policy cannot get past the lift phase. Object-in-scene LL training is the single most critical ingredient: the surrogate recovers only to 6.7\,\%, in stark contrast to the 77.1\,\% the same surrogate achieves in the comparison above. The reason is that our LL is a compact motion-imitation policy that, without the box, ends up imitating pickup motions with nothing to pick up, whereas the comparison WBC is a broad command-tracking controller that can be driven into a box-holding configuration regardless. Unlike the baselines, the ablation collapses are roughly uniform across light and heavy boxes: these failures are structural rather than load-dependent.

\begin{table}[t]
\centering
\captionsetup{width=0.84\columnwidth}
\caption{Ablation study (20\,s). Phase $1\!\rightarrow\!2$: \% of all episodes that complete the lift phase. OOD freq.: fraction of steps on which the pre-clip HL command lies outside the LL's training envelope. N\&S idx as in Sec.~\ref{sec:sim_compare}.}
\label{tab:sim_ablation}
\tablefont
\setlength{\tabcolsep}{5pt}
\rowcolors{2}{white}{gray!15}
\begin{tabular}{lcccccc}
\toprule
Variant & \makecell{Succ.\\(all)} & \makecell{Succ.\\(light)} & \makecell{Succ.\\(heavy)} & \makecell{Phase\\$1\!\rightarrow\!2$} & \makecell{OOD\\freq.} & \makecell{N\&S\\idx} \\
\midrule
\textbf{Ours (full)} & \textbf{80.7} & \textbf{85.8} & \textbf{70.5} & \textbf{91.7} & \textbf{0.15} & \textbf{1.00} \\
\midrule
w/o observation noise           & 3.0  & 2.7  & 3.6  & 38.6 & 0.45 & 2.05 \\
w/o object (no EE surrogate)    & 2.9  & 3.1  & 2.5  & 3.1  & 0.25 & 1.23 \\
w/o object (EE force surrogate) & 6.7  & 8.3  & 3.6  & 27.0 & 0.24 & 1.27 \\
w/o targeted HL RSI             & 6.5  & 6.6  & 6.4  & 17.5 & 0.19 & 1.05 \\
w/o dataset-envelope clipping   & 15.5 & 15.6 & 15.2 & 17.1 & 1.00 & 2.65 \\
\bottomrule
\end{tabular}
\end{table}

\vspace{-2pt}
\section{Limitations and Future Work}
\label{sec:limitations}

Three limitations frame the scope of this work and the natural next steps. \emph{(i) Perception.} The deployed system reads the 6-DoF box pose from an external motion-capture stream, so we have not validated the controller against fully on-board perception. The natural next step is to close the perception loop, for example by using our HL policy as a teacher and distilling it into a depth-based student, while the goal would be tracked in the robot frame via on-board localization. \emph{(ii) Motion-library coverage.} The four primitive families on which the LL motion prior is built limit downstream behaviors: table-to-table transfers beyond the demonstrated lift range, diverse initial grasps, or non-prehensile interactions such as pushing or tumbling, require additional retargeted demonstrations; the HL policy itself cannot extrapolate to them. Porting the framework to other contact-rich loco-manipulation tasks (door opening, non-prehensile object transport) is modest but not automatic: it requires new demonstrations, a phase graph for the new stages, and stage-driven rewards, while the rest of the LL and HL formulation carries over unchanged. \emph{(iii) Naturalness.} The N\&S index is only a proxy for naturalness; a proper user study would be needed to assess it.

\vspace{-2pt}
\section{Conclusion}
\label{sec:conclusion}

We presented a hierarchical RL framework for humanoid loco-manipulation in which a single-stage, object-in-scene LL motion prior is trained and then steered by a reference-free HL policy that requires only a sparse user-supplied goal. This setup absorbs the real distributed multi-contact wrench at its source and inherits natural human-like coordination from retargeted MoCap; the HL MDP is designed to drive the long-horizon task while structurally preserving the LL motion prior. On a Unitree~G1, a single deployed policy executes bimanual box transport across box geometries, payloads up to the platform's actuation envelope, multi-goal and multi-box sequences, and ground-to-table transfers, and reactively recovers from mid-carry pushes and table-edge collisions.

\ifx\paperformat\formatieee
    \bibliographystyle{IEEEtran}
\fi
\clearpage  %
\bibliography{references}  %

\clearpage
\appendix
\setlength{\textfloatsep}{7pt plus 2pt minus 2pt}
\setlength{\intextsep}{7pt plus 2pt minus 2pt}

\section{Motion Retargeting and Object Augmentation}
\label{app:retargeting}

To generate a dataset that generalizes across object sizes, we retarget
human MoCap data onto the Unitree~G1 humanoid using a
custom implementation of the interaction-mesh
formulation~\cite{ho2010interaction,OmniRetarget2025}. The first primitive family of
Sec.~\ref{sec:method:retargeting} (free locomotion) does not involve the box, so
its reference is reused directly from a ZEST~\cite{ZEST2025} retargeted random-walk clip
and does not enter the interaction-mesh pipeline. The remaining three primitive
families are seeded by three MoCap demonstrations of a subject interacting with
a box:
\begin{itemize}
    \item \textbf{carry:} random walk while holding the box at chest height.
    \item \textbf{pick:} bend over and pick up the box from the floor.
    \item \textbf{place:} while carrying the box, bend over and place
    it on the ground.
\end{itemize}
From these three primitives we produce 30 training clips through a
three-stage pipeline: preprocessing the MoCap data to match G1's anthropometry
(Section~\ref{sec:scaling}), running the interaction-mesh solver, and
re-solving against varying box geometries to augment the seed set
(Section~\ref{sec:augmentation}).

\subsection{Mocap Data Scaling}
\label{sec:scaling}

Since the interaction-mesh formulation preserves relative spatial
distances, it is sensitive to several mismatches between source and
target. It is sensitive to differences in limb proportions: G1's limb
ratios are not those of the human subject, so no single global scalar
can match every segment length at once. It is also sensitive to the
coupling between root displacement and leg-swing length: if the pelvis
traverses more or less ground than the feet can cover, foot-sliding
appears as a pre-retargeting artifact. The last issue is contact
preservation at the grasp, where the wrist-to-wrist span shrinks with
scaling and pushes the palms inside the object. We account for these by
(i)~splitting the root scale into a horizontal stature ratio
$s_{\text{height}} = h_\text{r}/h_\text{h}$ and a vertical leg-chain
ratio $s_{\text{leg}} = \ell_\text{r}/\ell_\text{h}$, so the pelvis
lands at G1's natural standing height while horizontal stride remains
proportional to overall body size; (ii)~rebuilding every kinematic
chain as $\mathbf{p}_{\text{child}}(t) =
\mathbf{p}_{\text{parent}}(t) + L_{\text{robot}}\,
\hat{\mathbf{d}}_{\text{human}}(t)$, with $L_{\text{robot}}$ taken from
neutral-pose forward kinematics and $\hat{\mathbf{d}}_{\text{human}}(t)$
taken from the \emph{unscaled} mocap so bone directions are not
distorted by the per-axis scaling; (iii)~regrounding per-frame against
the stance toe rather than a global $z$-min, which pins the supporting
foot to the floor through flight phases; and (iv)~freezing the hand-box
contact points and solving for wrist orientations that minimize the
deviation between palm normals and their original grasp axes.

\subsection{Object-Geometry Augmentation}
\label{sec:augmentation}

To avoid over-fitting to a single object, we re-instantiate each
retargeted trajectory with boxes of varying dimensions and re-solve the
mesh, in the spirit of the spatial augmentations of
\cite{OmniRetarget2025}. The source cube is expanded into nine variants per
behaviour (Table~\ref{tab:box-aug}), yielding 30 executable
demonstrations from a single MoCap recording.

\begin{table}[H]
\centering
\caption{Object-geometry augmentation. Dimensions in meters; each box is
instantiated for all three behaviours (\emph{carry}, \emph{pick},
\emph{place}).}
\label{tab:box-aug}
\setlength{\tabcolsep}{5pt}
\renewcommand{\arraystretch}{1.10}
\begin{tabular}{@{}lcc@{}}
\toprule
\textbf{Variant} & \textbf{Dimensions $(L\!\times\!W\!\times\!H)$} & \textbf{\#} \\
\midrule
Source       & $0.32\times0.32\times0.32$         & 1 \\
\midrule
Large box    & permutations of $(0.39,\,0.42,\,0.54)$ & 6 \\
Medium box   & permutations of $(0.32,\,0.32,\,0.46)$ & 3 \\
\midrule
\multicolumn{2}{l}{\textbf{Variants per behaviour}}   & \textbf{10} \\
\multicolumn{2}{l}{\textbf{Total demonstrations}}     & \textbf{30} \\
\bottomrule
\end{tabular}
\end{table}

\newpage
\section{Actor-Critic Observations}
\label{app:obs}

All actor terms have additive Gaussian noise with the per-dimension standard
deviations of Table~\ref{tab:actor-obs}; all critic terms are noise-free.

\begin{table}[H]
    \centering
    \caption{\textbf{HL actor (deployment) observations.}  Every term is
        available from on-board sensing or the mocap stream.  ``Permute''
        denotes that the six face-center entries are randomly permuted at
        every environment reset (see Sec.~\ref{sec:method:hl}).}
    \label{tab:actor-obs}
    \small
    \rowcolors{2}{white}{gray!15}
    \begin{tabular}{lcll}
        \toprule
        \textbf{Term} & \textbf{Dim} & \textbf{Source} & \textbf{Noise (Gaussian std)} \\
        \midrule
        Box face centers (torso frame, permute) & $18$ & Mocap & 0.10 m (clipped to $[-3,3]$) \\
        IMU base angular velocity               & $3$  & IMU @ torso & $0.10\,\text{rad/s}$ \\
        Projected gravity $\hat{g}_b$ (torso frame) & $3$  & IMU @ torso & $0.015$ \\
        Joint positions                         & $n_q$ & Encoders & $0.005\,\text{rad}$ \\
        Joint velocities                        & $n_q$ & Encoders & $0.25\,\text{rad/s}$ \\
        Last HL action $\mathbf a^{\text{HL}}_{t-1}$ & $12$ & Internal & --- \\
        Last LL action $\mathbf a^{\text{LL}}_{t-1}$ & $n_q$ & Internal & --- \\
        Goal in torso frame $\mathbf g$         & $3$  & User input & $0.10\,\text{m}$ (clipped to $[-3,3]$) \\
        \bottomrule
    \end{tabular}
\end{table}

\begin{table}[H]
    \centering
    \caption{\textbf{HL critic (training-only, privileged) observations}
        added on top of the actor observations of Table~\ref{tab:actor-obs}.
        All terms are read directly from the simulator, are noise-free, and
        are never accessed at deployment.}
    \label{tab:critic-obs}
    \small
    \rowcolors{2}{white}{gray!15}
    \begin{tabular}{ll}
        \toprule
        \textbf{Category} & \textbf{Term} \\
        \midrule
        Ground-truth kinematics
            & Robot/object root position \\
            & Robot/object linear \& angular velocity \\
            & Robot/object projected gravity \\
            & Robot key-point body positions/velocities \\
            & Closest box point (torso frame) \\
        \midrule
        Randomized parameters
            & Robot/object per-body mass \\
            & Robot/object material properties (friction, restitution, \ldots) \\
        \midrule
        Contact
            & Wrist contact forces \\
            & Box contact forces \\
        \midrule
        Scheduler & Task-phase one-hot $s_t$ \\
        \bottomrule
    \end{tabular}
\end{table}

\paragraph{LL observations}
The LL policy observes proprioception only (IMU angular velocity,
projected gravity, joint positions, joint velocities, and its own previous
action) plus the 12-D command $\mathbf c^{\text{LL}}_t$.  The same Gaussian
noise as in Table~\ref{tab:actor-obs} is applied during LL training to its
proprioceptive terms; in addition, \emph{bounded uniform} noise is injected on
each component of $\mathbf c^{\text{LL}}_t$, with per-dimension bounds chosen
to reflect reasonable tracking tolerances: $\pm 0.10\,\text{m}$ on base height,
$\pm 0.15\,\text{rad}$ on base orientation, $\pm 0.25\,\text{m/s}$ on the
planar base twist, and $\pm 0.05\,\text{m}$ on each end-effector position.
We use uniform rather than Gaussian noise to keep the perturbed command
inside a bounded box around the reference; empirically, replacing it with
matched-scale Gaussian noise yields comparable performance, and both
substantially outperform no command noise.

\section{Residual-removal curriculum ablations}\label{app:residual_curriculum}

\begin{table}[H]
\centering
\caption{Residual-removal curriculum on LL training. Survival: fraction of environments not terminated at episode end. Tracking similarity: motion-imitation similarity score between the policy rollout and the retargeted reference. All variants trained for the same number of iterations.}
\label{tab:residual_curriculum}
\rowcolors{2}{white}{gray!15}
\begin{tabular}{lcc}
\toprule
Feedforward baseline & Survival (\%) & Tracking similarity (\%) \\
\midrule
Curriculum $\beta_t: 0 \to 1$ (ours) & \textbf{82} & \textbf{89} \\
Fixed default joint positions & 72 & 83 \\
Fixed measured joint positions & 73 & 85 \\
\bottomrule
\end{tabular}
\end{table}

\section{Task Phase Scheduler}
\label{app:planner}

Formally, the scheduler is a deterministic finite-state machine
$\mathcal{M} = (\mathcal{S}, s_0, \delta)$ with state set
$\mathcal{S} = \{0,1,2,3,4\}$, comprising four active phases
($s_t \in \{0,1,2,3\}$) and a single absorbing \emph{idle} index
($s_t = 4$), and a transition function
$\delta : \mathcal{S}\times\Omega \to \mathcal{S}$, where $\Omega$ is the
physical state of the environment.  Each forward transition is gated by a
Boolean predicate $g_s : \Omega \to \{0,1\}$ over physically meaningful
quantities (position and heading tolerances, bilateral arm--box contact
$c$, and box height); phase~2 additionally admits a drop-fallback
predicate $g_{\text{drop}}$, and $g_{\text{done}}$ denotes the placement
predicate.  The transition
function is
\begin{equation}
\delta(s,\omega) =
\begin{cases}
4 & \text{if } g_{\text{done}}(\omega), \\
0 & \text{if } s = 2 \,\wedge\, g_{\text{drop}}(\omega), \\
s+1 & \text{if } s \in \{0,1,2,3\} \,\wedge\, g_s(\omega), \\
s & \text{otherwise,}
\end{cases}
\label{eq:fsm}
\end{equation}
with initial state $s_0(\omega) = 4$ if $g_{\text{done}}(\omega)$ at reset
and $s_0(\omega) = 0$ otherwise.  Active-phase transitions are
summarized in Table~\ref{tab:planner}.  An episode normally proceeds
$0\!\to\!1\!\to\!2\!\to\!3$; the carry phase additionally admits a
\emph{drop} fallback to phase~0 when the lowest box-contact height falls
below a release threshold $z_{\text{drop}}$ or bilateral contact $c$ is
lost.  When the placement predicate fires (box xy and lowest-contact height
both within tolerance of $\mathbf{g}$), the scheduler enters the absorbing
idle index in which the policy is rewarded for stopping locomotion and
releasing the box (reward $R_4$, Eq.~\ref{eq:r4}); the same predicate is
also evaluated at reset, so an episode whose initial box pose is already
within tolerance starts directly in idle.  Recall that the scheduler is
used \emph{only at training time}, to (a)~supply the per-phase reward
index $s_t$ and (b)~provide the critic with privileged phase information;
the deployed actor never observes $s_t$.

\begin{table}[H]
    \centering
    \caption{\textbf{Four-phase task scheduler (active phases).}  $\mathbf p_o^{\text{face}}$
        denotes the closest of the six box face centers to the robot torso,
        and $\mathbf p_o$ denotes the box center;  $\theta_e$
        denotes the heading error from the robot torso to the box
        (phase~0) or to the goal (phase~2); $c$ denotes the binary event
        ``both wrists in contact with the box.''  Phase~2 additionally has
        a drop edge back to phase~0 (carry-to-goal $\to$ walk-to-box) when
        the lowest box-contact height falls below $z_{\text{drop}}$ or $c$
        is lost.  When the placement predicate fires the scheduler enters
        an absorbing idle index (omitted from the table).  Numerical values
        are the defaults used in our experiments.}
    \label{tab:planner}
    \small
    \setlength{\tabcolsep}{4pt}
    \rowcolors{2}{white}{gray!15}
    \begin{tabular}{@{}cllp{0.50\linewidth}@{}}
        \toprule
        $s_t$ & \textbf{Phase} & \textbf{Goal} & \textbf{Transition trigger} \\
        \midrule
        0 & \textsc{walk-to-box}    & approach the box               & $\|\mathbf p_b - \mathbf p_o^{\text{face}}\|_{xy} < 0.50\,\text{m}$ and $|\theta_e|<0.40$ \\
        1 & \textsc{lift-box}       & bilateral grasp \& lift        & $c$ and lowest box-contact height $\geq 0.65\,\text{m}$ \\
        2 & \textsc{carry-to-goal}  & carry box to $\mathbf g$       & $\|\mathbf p_o - \mathbf g\|_{xy} < 0.50\,\text{m}$ and $|\theta_e|<0.40$ \\
        3 & \textsc{place-box}      & lower the box onto $\mathbf g$ & box xy within $0.50\,\text{m}$ and lowest-contact height within $0.15\,\text{m}$ of $z^\star$ \\
        \bottomrule
    \end{tabular}
\end{table}

\section{Per-phase reward terms}\label{app:reward_terms}

With
\begin{equation*}
f_v(x) = \tanh(\alpha_v x), \qquad f_\sigma(\xi) = \exp\!\bigl(-\|\xi\|^2/(2\sigma^2)\bigr),
\end{equation*}
where $\alpha_v$ denotes one of two gains depending on whether its argument is a robot- or box-velocity quantity, the per-phase rewards $R_s$ entering Eq.~\ref{eq:task_reward} are
\begin{align}
R_0 &= \tfrac{1}{2}\bigl[f_v(\mathbf{v}^{xy}_b\cdot\hat{\mathbf{e}}^{xy}_{bo}) + f_\sigma(\theta_e)\bigr], \label{eq:r0} \\[6pt]
R_1 &= \tfrac{1}{2}\mathbf{1}_{\{\neg c\}}\bigl[f_v(\mathbf{v}_r\cdot\hat{\mathbf{e}}_r) + f_v(\mathbf{v}_l\cdot\hat{\mathbf{e}}_l)\bigr] \nonumber\\
    & \ + \tfrac{1}{2}\mathbf{1}_{\{c\}}\bigl[f_v(\mathbf{v}^z_o\cdot\hat{\mathbf{e}}_o^z) + \mathbf{1}_{\{c\}}\bigr], \\[6pt]
R_2 &= \tfrac{1}{3}\bigl[f_v(\mathbf{v}_o^{xy}\cdot\hat{\mathbf{e}}_{bg}^{xy}) + f_\sigma(\theta_e) + \mathbf{1}_{\{c\}}\bigr], \\[6pt]
R_3 &= \tfrac{1}{2}\bigl[f_v(\mathbf{v}_o\cdot\hat{\mathbf{e}}_{bg}) + \mathbf{1}_{\{c\}}\bigr], \\[6pt]
R_4 &= \tfrac{1}{3}\bigl[f_\sigma(\mathbf{v}_b) + f_\sigma(\boldsymbol{\omega}_b) + \mathbf{1}_{\{\neg c\}}\bigr], \label{eq:r4}
\end{align}
where all vector quantities are expressed in the world frame, and the superscript ${xy}$ denotes projection onto the horizontal plane. $\mathbf{v}_b$ and $\boldsymbol{\omega}_b$ are the robot base linear and angular velocities, $\mathbf{v}_o$ is the box velocity. The unit vectors $\hat{\mathbf{e}}_{bo}$ and $\hat{\mathbf{e}}_{bg}$ point from the robot's base to the closest box face center and to the goal respectively, while $\hat{\mathbf{e}}_o$ points from the box center toward the target carry height. For the end-effectors, $\mathbf{v}_r$ and $\mathbf{v}_l$ are the right and left velocities, and $\hat{\mathbf{e}}_r$, $\hat{\mathbf{e}}_l$ point toward the nearest pair of opposing side faces on the box. The heading error $\theta_e$ is measured against $\hat{\mathbf{e}}_{\text{head}} = -\hat{\mathbf{n}}_{\text{approach}}$, orienting the robot to face the approach side, and $\mathbf{1}_{\{c\}}$ is the indicator function for bilateral end-effector contact with the box.

\section{Reward Weights}
\label{app:reward}

The HL training objective is a weighted sum of the bounded staged reward of
Sec.~\ref{sec:hl_reward} and standard
regularizers.  The complete weight
table is given in Table~\ref{tab:reward-weights}.

\begin{table}[H]
    \centering
    \caption{\textbf{HL reward weights} used during training.  Positive
        weights reward desired behavior; negative weights penalize deviations.}
    \label{tab:reward-weights}
    \small
    \setlength{\tabcolsep}{4pt}
    \rowcolors{2}{white}{gray!15}
    \begin{tabular}{lcl}
        \toprule
        \textbf{Term} & \textbf{Weight} & \textbf{Description} \\
        \midrule
        Bounded staged reward $R_{s_t}$        & $+10.0$                & Phase-conditional task reward, Eqs.~(\ref{eq:r0})--(\ref{eq:r4}) \\
        Alive bonus                            & $+5.0$                 & Constant per-step bonus discouraging early termination \\
        Action clip penalty                    & $-10.0$                & Penalty on pre-clipped HL actions that violate command limits \\
        HL action smoothness                   & $-0.10$                & Penalty on $\|\mathbf a^{\text{HL}}_t - \mathbf a^{\text{HL}}_{t-1}\|^2$ \\
        LL action smoothness                   & $-0.25$                & Penalty on the realized LL action rate \\
        Joint torques                          & $-5\!\times\!10^{-4}$  & Penalty on squared joint-torque magnitudes \\
        DoF position-limit violation           & $-10.0$                & Penalty for proximity to joint position limits \\
        Applied torque limit                   & $-1.0$                 & Penalty on saturated joint torques \\
        Illegal contact (large force)          & $-10.0$                & Penalty on contact forces above $6000\,\text{N}$ on any link \\
        Foot slip                              & $-0.5$                 & Penalty on tangential foot velocity at contact \\
        HL/LL state mismatch                   & $-1.0$                 & Penalty on tracking residual of $\mathbf c^{\text{LL}}_t$ \\
        \bottomrule
    \end{tabular}
\end{table}

\section{PPO Hyperparameters}\label{app:training_setup}

\begin{table}[H]
    \centering
    \caption{\textbf{PPO hyperparameters} for the LL and HL
        policies.}
    \label{tab:ppo}
    \small
    \rowcolors{2}{white}{gray!15}
    \begin{tabular}{ll}
        \toprule
        \textbf{Hyperparameter} & \textbf{Value} \\
        \midrule
        Actor network                              & MLP$[512, 256, 128]$ with ELU activation \\
        Critic network                             & MLP$[512, 512, 256]$ with ELU activation \\
        Parallel environments $N_\text{env}$       & $8192$ \\
        Steps per env per update $T$               & $24$ \\
        Max iterations                             & $30{,}000$ \\
        Discount $\gamma$                          & $0.99$ \\
        GAE $\lambda$                              & $0.95$ \\
        Clip range                                 & $0.2$ \\
        Value-loss coefficient                     & $0.5$ \\
        Entropy coefficient                        & $0.001$ \\
        Learning epochs per update                 & $5$ \\
        Mini-batches per epoch                     & $4$ \\
        Learning rate                              & $1\!\times\!10^{-3}$ (adaptive) \\
        Target KL                                  & $0.01$ \\
        Max gradient norm                          & $1.0$ \\
        Empirical observation normalization        & on \\
        Clipped value loss                         & on \\
        \bottomrule
    \end{tabular}
\end{table}

\section{Domain Randomization}
\label{app:dr}

Table~\ref{tab:dr} lists the randomization ranges applied when spawning our environments or at the start of 
each episode (\textit{startup}/\textit{reset}), together with the impulsive
disturbances applied periodically during the episode (\textit{interval}). Across all episodes, the
box is sampled from three fixed geometries used during training---small
$(0.32, 0.32, 0.32)\,\text{m}$, medium $(0.46, 0.32, 0.32)\,\text{m}$, and
large $(0.54, 0.39, 0.42)\,\text{m}$ edge lengths $(L_x, L_y, L_z)$.

\paragraph{Source- and target-side support tables}
Although the LL motion library contains only ground-level pickup and placement primitives, the HL training environment additionally spawns a support table at the target location with probability $0.5$, at heights up to $75$~cm---the upper end of the carry-height range covered by the LL pickup and placement references. Source-side variation is more conservative, with initial heights up to $50$~cm and a lower probability of $0.15$: since the HL does not observe the table, a collision with a source-side table edge during the approach or grasp would compromise the rest of the episode, whereas a comparable collision at placement occurs once the box is already near its goal and is therefore far less costly. Such collisions are treated as external disturbances the policy learns to handle. This extension tests whether the HL policy can adapt ground-trained LL primitives to elevated targets without retraining the LL on additional references.

\begin{table}[H]
    \centering
    \caption{\textbf{Domain randomization} used during HL training.  The LL
        stage uses the same ranges except that (i) no source/target table is
        present, (ii) no goal is sampled, and (iii) the LL always uses RSI
        for initialization.  Distributions are uniform over the indicated
        intervals unless stated otherwise. Rows marked ($\dagger$) apply
        only on resets where RSI is not active.}
    \label{tab:dr}
    \small
    \rowcolors{2}{white}{gray!15}
    \begin{tabular}{lll}
        \toprule
        \textbf{Quantity} & \textbf{Range} & \textbf{Mode} \\
        \midrule
        Robot static friction                 & $[0.6,\,1.2]$              & startup \\
        Robot dynamic friction                & $[0.5,\,1.1]$              & startup \\
        Robot restitution                     & $[0.0,\,0.4]$              & startup \\
        Robot per-body mass scale             & $[0.9,\,1.1]$            & startup \\
        Robot torso mass scale                & $[0.8,\,1.2]$            & startup \\
        Actuator stiffness scale              & $[0.8,\,1.2]$              & startup \\
        Actuator damping scale                & $[0.8,\,1.2]$              & startup \\
        Object static friction                & $[0.4,\,1.2]$              & startup \\
        Object dynamic friction               & $[0.3,\,1.1]$              & startup \\
        Object restitution                    & $[0.0,\,0.4]$              & startup \\
        Object mass scale                     & $[0.5,\,5.0]$              & startup \\
        Source-table top height               & $[0.3,\,0.5]\,\text{m}$  & reset \\
        Source-table absent (no source table) & with prob.\ $0.85$         & reset \\
        Target-table top height               & $[0.3,\,0.75]\,\text{m}$  & reset \\
        Target-table absent (place on ground) & with prob.\ $0.50$         & reset \\
        Robot reset $x$ (planar offset)$^\dagger$       & $[-6.0,\,0.0]\,\text{m}$   & reset \\
        Robot reset yaw$^\dagger$                       & $[-\pi,\,\pi]\,\text{rad}$ & reset \\
        Robot reset joint perturbation        & $\pm 0.2\,\text{rad}$, $\pm 0.5\,\text{rad/s}$ & reset \\
        Object reset $x,y$ offset             & $[-0.20,\,0.20]\,\text{m}$ & reset \\
        Object reset yaw$^\dagger$                      & $[-\pi,\,\pi]\,\text{rad}$ & reset \\
        HL RSI probability                    & $0.75$                     & reset \\
        Robot impulsive push (linear)         & $\pm 0.5\,\text{m/s}$ ($x,y$), $\pm 0.2$ ($z$) & every $0$--$5\,\text{s}$ \\
        Object impulsive push (linear)        & $\pm 0.3\,\text{m/s}$ ($x,y$) & every $0$--$5\,\text{s}$ \\
        \bottomrule
    \end{tabular}
\end{table}

\section{Joint-torque trace on heavy payload}\label{app:torque_trace}

Figure~\ref{fig:torque_heatmap} shows the per-joint torque trace for the 7~kg hardware pick-and-place, normalized by each joint's effort limit, alongside snapshots of the lift and carry phases. Three joints exceed their rating during the sustained high-load epoch ($t\!\approx\!4$--$15$~s): the waist-pitch joint peaks at 116.2\,\% of its limit ($58.1$~Nm / $50$~Nm) at $t\!\approx\!4.6$~s under the lifting load; the right ankle-pitch joint reaches 114.2\,\% ($57.1$~Nm / $50$~Nm) at $t\!\approx\!14.3$~s while stabilizing the shifted center of mass; and the left elbow touches 107.3\,\% ($26.8$~Nm / $25$~Nm) at $t\!\approx\!9.1$~s during the lift. Several more joints operate in the 80--99\,\% band: left ankle-pitch (98.8\,\%), right elbow (97.5\,\%), right shoulder-pitch (92.6\,\%), left shoulder-pitch (87.1\,\%), and left hip-pitch (82.5\,\%). The pattern matches the task: the elbows and shoulders bear the lifted payload while the ankles, waist, and hips absorb the shifted center of mass through the carry phase.

\section{Naturalness-and-Smoothness Index}
\label{app:ns_index}

The N\&S index used in Section~\ref{sec:sim_compare} is the unweighted average of six lower-is-better motion-quality ratios, each computed as the method's value divided by the value of Ours on the same metric:
\begin{itemize}
    \item \emph{OOD-command frequency}: the fraction of control steps on which the HL's pre-clip command lies outside the envelope the LL was trained on (the no-clipping ablation also removes the pre-clip penalty).
    \item \emph{Posture deviation}: deviation of the base height, roll, and pitch from the mean reference during walking and carrying.
    \item \emph{Arm-pose deviation}: deviation of left and right end-effector positions (torso frame) from the mean reference during walking and carrying.
    \item \emph{HL action smoothness}: $\|\mathbf{a}^{\text{HL}}_t - \mathbf{a}^{\text{HL}}_{t-1}\|^2$ averaged over the rollout.
    \item \emph{LL action smoothness}: $\|\mathbf{a}^{\text{LL}}_t - \mathbf{a}^{\text{LL}}_{t-1}\|^2$ averaged over the rollout.
    \item \emph{Joint effort}: $\|\boldsymbol{\tau}_t\|^2$ averaged over the rollout.
\end{itemize}
By construction, Ours has N\&S$=1.00$; a value above $1.00$ for a baseline indicates motion that is, on average, proportionally less smooth, more effortful, or further from the reference than Ours.

\section{Per-Phase Failure Breakdown}
\label{app:phase}

\begin{table}[H]
    \centering
    \caption{\textbf{Per-phase completion}: \% of all episodes that pass each
        scheduler transition (Appendix~\ref{app:planner}) at least once (20\,s
        episode length unless noted). The last column equals task success, so the drop between
        consecutive columns gives the failures in each phase.}
    \label{tab:phase}
    \small
    \rowcolors{3}{gray!15}{white}
    \begin{tabular}{lcccc}
        \toprule
        \textbf{Setting} & \textbf{Reached box} & \textbf{Lifted box} & \textbf{Reached goal} & \textbf{Placed (succ.)} \\
        & ($0\!\rightarrow\!1$) & ($1\!\rightarrow\!2$) & ($2\!\rightarrow\!3$) & ($3\!\rightarrow\!4$) \\
        \midrule
        Ours                   & 97.1 & 91.7 & 82.1 & 80.7 \\
        Ours, 40\,s            & 96.9 & 92.3 & 88.0 & 87.9 \\
        \midrule
        WBC (no object)        & 96.8 & 85.0 & 53.5 & 50.3 \\
        WBC (force surrogate)  & 98.2 & 91.7 & 77.3 & 77.1 \\
        \bottomrule
    \end{tabular}
\end{table}

\section{Robustness Analysis}
\label{app:robustness}

\begin{figure}[H]
\centering
\safeincludegraphics[width=\columnwidth]{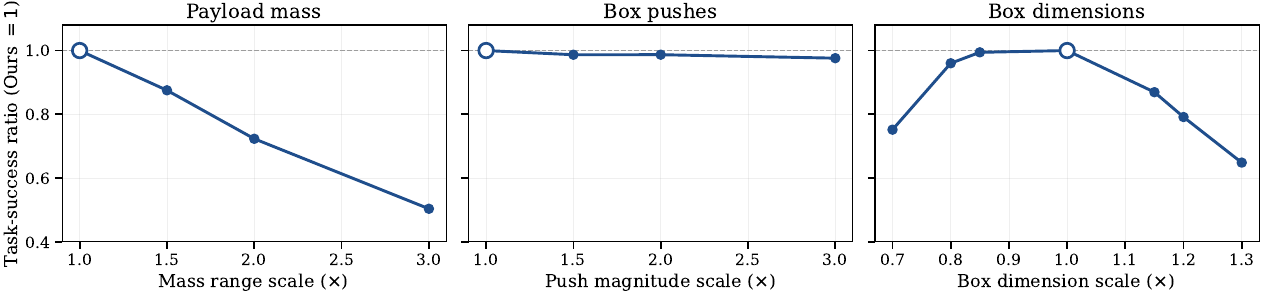}
\caption{Robustness Study (20\,s). Task-success ratio (Ours nominal $=1$) under 3 perturbation axes (box mass, external pushes, box dimensions) scaled relative to the training distribution.}
\label{fig:sim_robust}
\end{figure}

Figure~\ref{fig:sim_robust} stress-tests the deployed policy along three axes, each scaled well beyond its training range; the $y$-axis is task success normalized so that the nominal Ours-20\,s policy sits at $1$. The policy is \emph{least} sensitive to external box pushes: even at $3\times$ the training disturbance magnitude, success holds at $0.98$ of nominal. Stronger pushes mainly translate into more frequent box drops rather than outright failures: the dropped-box rate rises only modestly, from 26.8\,\% at nominal to 29.1\,\% at $3\times$, and the policy could still re-acquire the dropped box and complete the task within the episode budget. It is \emph{moderately} sensitive to payload mass, degrading smoothly to $0.50$ of nominal when the mass range is tripled (mean payload 10.3\,kg); the bottleneck here is raw payload against the platform's actuation limits. The policy is \emph{most} sensitive to box dimensions: $\pm15\,\%$ is graceful ($0.87$--$0.99$ of nominal), but $\pm30\,\%$ drops success to $0.65$--$0.75$, since box geometry changes the grasp and contact configuration that the face-center observation can only partially absorb. The overall sensitivity ordering is dimensions $>$ mass $\gg$ pushes.

\end{document}